\documentclass[10pt,conference]{ieeeconf} % uncomment this

\usepackage{comment}
\usepackage{setspace}
\usepackage{amssymb, amsmath, cite}
\usepackage{color}
\usepackage{verbatim}
\usepackage{mathrsfs}
\usepackage{color,hyperref}
\definecolor{darkblue}{rgb}{0.0,0.0,0.5}
\hypersetup{colorlinks,breaklinks,
            urlcolor=darkblue,
            citecolor=darkblue,
            linkcolor=darkblue}
            
\usepackage{stmaryrd}
\usepackage[top=1in, bottom=0.7in, left=0.7in, right=0.7in]{geometry}
\usepackage{mathtools}
\usepackage{url}
\usepackage{float}

\let\labelindent\relax
\usepackage{enumitem}
\setlist[description]{leftmargin=0.3cm,labelindent=0cm}

\newtheorem{theorem}{Theorem}[section]

\newtheorem{defn}{Definition}[section]

\newtheorem{problem}{Problem}[section]
\newtheorem{example}{Example}[section]

\IEEEoverridecommandlockouts % uncomment this
\newcommand{\RR}{{\mathbb R}}

\newcommand{\ZZ}{{\mathbb Z}}

\newcommand{\cB}{{\mathcal B}}

\newcommand{\cF}{{\mathcal F}}

\newcommand{\cH}{{\mathcal H}}

\newcommand{\cL}{{\mathcal L}}

\newcommand{\cP}{{\mathcal P}}

\newcommand{\cX}{{\mathcal X}}

\newcommand{\Ys}{Y^{\star}}

\newcommand{\sH}{\mathscr{H}}
\newcommand{\sF}{\mathscr{F}}
\newcommand{\sB}{\mathscr{B}}
\newcommand{\sHs}{\mathscr{H}^{\star}}
\newcommand{\sFs}{\mathscr{F}^{\star}}
\newcommand{\sBs}{\mathscr{B}^{\star}}

\newcommand\tqed{\leavevmode\unskip\penalty9999 \hbox{}\nobreak\hfill\quad\hbox{$\triangleleft$}}

\title{\Large \bf
Hierarchically Consistent Motion Primitives for Quadrotor Coordination}

\author{Marijan Vukosavljev, Angela P. Schoellig, and Mireille E. Broucke% <-this % stops a space
\thanks{Marijan Vukosavljev and Mireille E. Broucke are with the Dept. of Electrical and Computer Engineering, 
University of Toronto, Canada (e-mails: mario.vukosavljev@mail.utoronto.ca,
broucke@control.utoronto.ca). Angela P. Schoellig is with the University of Toronto Institute for 
Aerospace Studies (UTIAS), Canada (email: schoellig@utias.utoronto.ca). Supported by the Natural Sciences 
and Engineering Research Council of Canada (NSERC). }%
}

\begin{document}
%\linespread{0.9} % in case tight for space!
\maketitle
\begin{abstract}
We present a hierarchical framework for motion planning of a large collection of 
agents. The proposed framework starts from low level motion primitives over a gridded 
workspace and provides a set of rules for constructing higher level 
motion primitives. Our hierarchical approach is highly scalable and robust making it an ideal tool for planning for multi-agent systems. Results are demonstrated experimentally on a collection of quadrotors that must navigate a cluttered environment while maintaining a formation.
\end{abstract}

\section{Introduction}

This paper proposes a hierarchical construction of motion primitives for motion planning of multi-vehicle systems. 
The framework allows designers to incrementally build up more complex motion primitives from simpler ones in a systematic, 
rigorous way, without the need to redesign feedback controllers. This approach dramatically reduces the overall complexity 
of motion planning. 

\begin{figure}%[t]
\centering%
\includegraphics[width=0.8\linewidth,trim=0cm 9.2cm 0cm 0cm, clip=true]{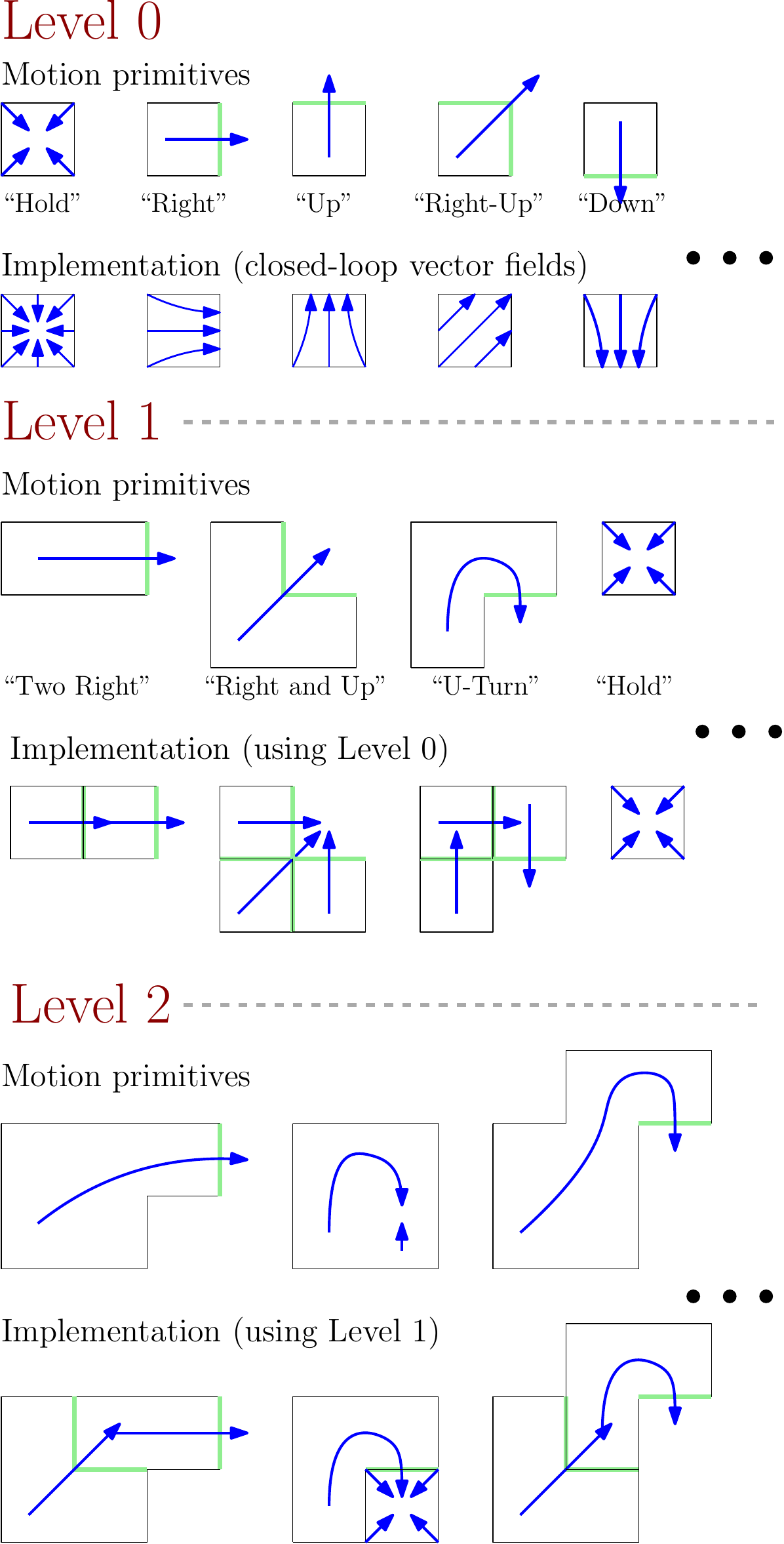}
\caption{An example of motion primitives designed at two hierarchical levels over a gridded two dimensional output space.} 
\vspace{-2mm}
\label{fig:motiveex}%
\end{figure}

The main idea is simple. Consider a control system with two outputs. Suppose the output space has been gridded into boxes, and 
we have feedback controllers for a finite set of {\em atomic motion primitives}. In Figure~\ref{fig:motiveex} we show five 
atomic motion primitives. Motion primitives at level $1$ are formed from sets of sequences of Level $0$ motion primitives. 
For example, two successive $Right$ atomic motion primitives from level $0$ form the $Two Right$ motion primitive at level $1$. 
Progressively more complex behaviors can be similarly defined, as shown in Figure~\ref{fig:motiveex}.
%Progressively more complex behaviors are similarly defined without compromising consistency with the vehicle dynamics.

Hierarchy is a common theme to simplify and scale up control design and has many applications \cite{KHAT05, PER11, KUL12, UDE15}. Motion primitives often serve as an important layer in the hierarchy and have been applied in various forms \cite{FRAZ05, KEL11, FARH14}. Abstraction as it relates to hierarchy has been studied in more general settings \cite{PAP00, PAP01, CAIN98, CAIN02, WON15}, but can be difficult to apply to practical applications.
One of our motivating applications is efficient formation coordination of quadrotors, which boasts a broad literature \cite{EGER01, AYAN10, SYCA17, BEL07, CASS16, KUM14, AHN15}.
%, multi-level hierarchical designs have not been investigated prominently.
%One of our motivating applications is efficient formation coordination of quadrotors. While there exists a broad literature on this topic, \cite{EGER01, AYAN10, SYCA17, BEL07, CASS16, KUM14, AHN15}, our approach offers a unique perspective that is both safe by design and computationally scalable in the number of vehicles.

Though our main idea is simple, and while many concepts and methods involving hierarchy, abstractions, and quadrotor control have been well studied, we place extra demands on our design that take it a step beyond what has been done. 
%The first additional  constraint is to 
First, we allow for an arbitrary number of hierarchical levels, and the design of any level depends in the same way only on the level below. 
%As such, motion primitives at any level can be used for synthesis of high-level motion planning objectives. % not clear
One of the benefits of this uniformity of the architecture is that a designer can apply a planning algorithm at any level to obtain a control synthesis to the problem.
Second, motion primitives at any level must be implementable by the low level continuous dynamics, for instance by adhering to safety and continuity constraints on positions and velocities. Effectively, we are implementing a notion of hierarchical consistency, but with the addition of a continuous state feedback at the lowest level \cite{CAIN02}. As a final constraint, we expect a design that gives dramatic computational advantages, especially in the multi-agent setting. To this end, we construct hierarchical motion primitives on a grid, as shown in Figure~\ref{fig:motiveex}.
%Even though the concept of hierarchy is often an intuitive one, in this paper we have placed some demands on the design and encoding of motion primitives. Our main goal was to define a suitable structure of hierarchy so that an arbitrary number of levels could be defined, and so that all levels have the same structure by depending recursively only on the level below. In this way, significant reduction in complexity can be achieved by planning only at the highest level of available motion primitives, which is guaranteed to be correctly implemented by the predefined design of the constituent motion primitives at the levels below. 
%Moreover, some interesting constraints on the design of motion primitives have been exposed, such as ...
%Moreover, an important feature of our hierarchical approach is that we incorporate continuous state feedback control at the lowest level along with the discrete structure of motion primitives at all the levels above so that the motion primitives can be applied to real robotic systems.

There are three main contributions of this paper. 
First, we provide an architectural formulation of hierarchical motion primitives, which involves careful attention to suitable, parsimonious data structures and relationships.
%The first is the architectural formulation of hierarchical motion primitives, which involved some careful attention to detail in determining suitable data structures and relationships among hierarchical levels. 
Second, we solve a reach-avoid problem with {\em behavior constraints},
%The second is the use of our hierarchical motion primitives to solve the reach avoid problem with {\em behavior constraints}, 
in which the vehicles must safely reach goal locations while maintaining predefined sequences of motions; here our analysis employs fairly standard notions from hybrid control theory. 
Finally, we supply an elegant and highly efficient design to address the formation control of quadrotors.
%Finally, within the proposed theoretical framework we have supplied a succinct and effective design to address the formation control of quadrotors. 
For example, we are able to navigate a formation of quadrotors interleaving them around a set of obstacles, while maintaining strict safety guarantees. 

%What distinguishes our hierarchical approach is that we span continuous state feedback control at the low level up to discrete planning at the higher levels, without resorting to approximations, while maintaining hierarchical consistency, and without compromising provably correct behavior. 

\begin{figure}%[t]
\centering%
\includegraphics[width=0.95\linewidth,trim=0cm 0cm 0cm 0cm, clip=true]{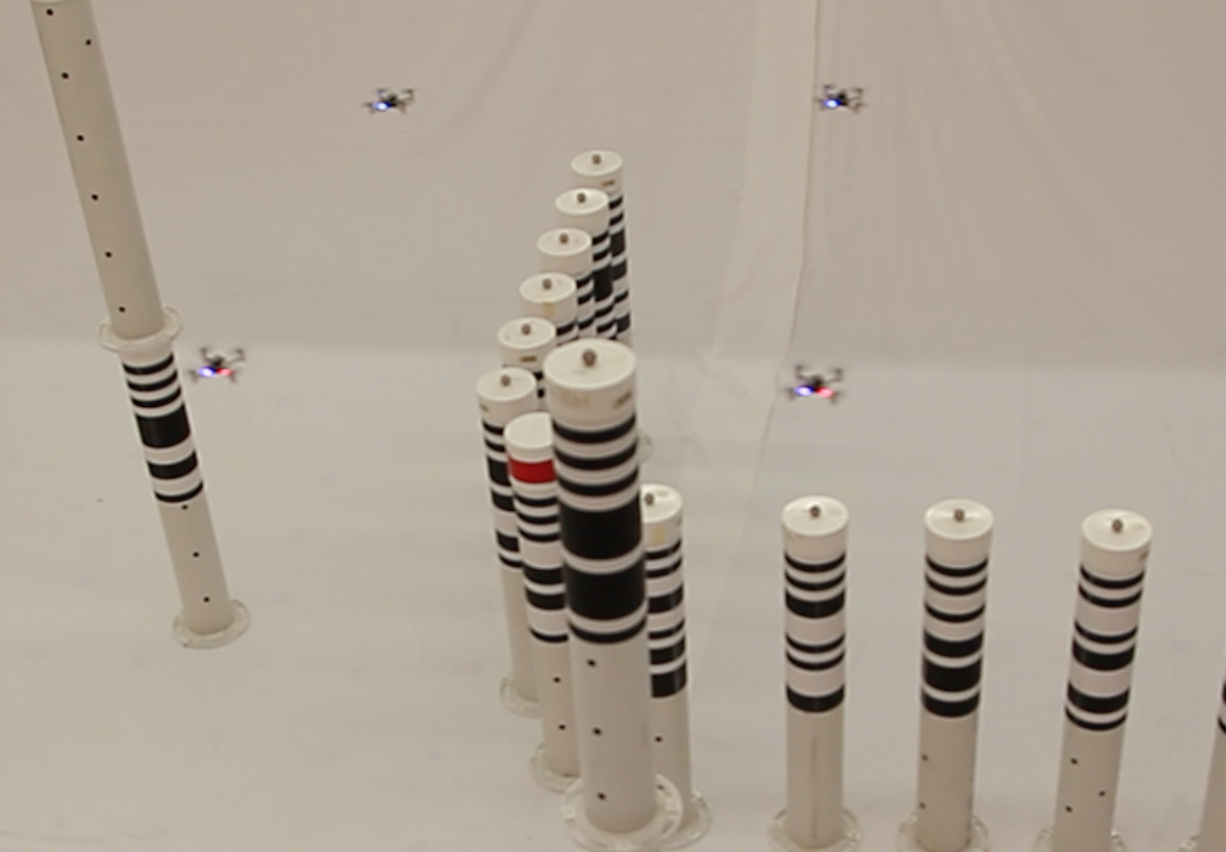}
\caption{Four tiny Crazyflie quadrotors navigate a cluttered environment while maintaining a square formation. A video illustrating the results can be found at http://tiny.cc/hier-moprim.} 
\vspace{-2mm}
\label{fig:exp_setup}%
\end{figure}

%A motivating application is coordination of quadrotors \cite{EGER01, AYAN10, SYCA17, BEL07, CASS16, KUM14, AHN15}. We formulate the formation control problem as a reach-avoid problem with {\em behavior constraints}. As far as we know, no other approach uses this formulation, and consequently we recover new behaviors in formation control. 

This research is an outgrowth of our modular framework for motion planning based on motion primitives \cite{VUK17,VUK18}; 
the notions of atomic motion primitives and the {\em maneuver automaton} were introduced there. As those constructions were limited only to level 0, this paper concerns itself with describing how to build hierarchical motion primitives at higher levels and illustrating the computational advantages of employing a multi-level hierarchy. Moreover, we have introduced the notion of behavior constraints in addition to the usual reach-avoid problem as a mechanism to address the formation control problem of quadrotors in a hierarchical manner.
At the same time we suppress theoretical details including proofs, which will be made available in a forthcoming journal paper. 

{\bf Notation}. Let $| \cdot |$ denote the cardinality of a set. If $A$ is a set, let $\cP(A)$ denote its power set. The set difference of $A$ and $B$ is denoted $A \setminus B$. For a collection of sets $\{A_i\}_{i = 1}^n$, the cartesian product is denoted $\prod_{i=1}^n A_i$; when $n = 2$, we may write $A_1 \times A_2$, and when $A_i = A$ for all $i =1, \ldots, n$, we may write $A^n$ (which in general should not be confused with superscripts for indexing in other contexts). Let $\ZZ$ and $\RR$ denote the integers and the real numbers, respectively.

\section{Problem Statement}
\label{sec:prob}

Consider the general nonlinear control system
\begin{equation}
\label{eq:thesystem}
\dot{x} = f(x,u) \,, \qquad \qquad y = h(x) \,,
\end{equation}
where $x \in \RR^n$ is the state, $u \in \RR^\mu$ is the input, and $y \in \RR^p$ is the output. Let  $y(\cdot,x_0)$ 
denote the output trajectory of \eqref{eq:thesystem} starting at initial condition $x_0 \in \RR^n$ and under some 
open-loop or feedback control. We make a standing assumption that the outputs of \eqref{eq:thesystem} are a subset 
of the states and that the vector field is invariant to the value of the output. This symmetry property is satisfied 
for many robotic systems, for example when the outputs are positions, and it allows motion primitives to be designed 
only over a single box in a gridded output space \cite{FRAZ05,VUK18}. 

Fix a {\em grid length vector} $d = (d_1, \ldots, d_p)$ where $d_i > 0$. The gridded output space is constructed 
by translating the canonical box $\Ys := \prod_{i=1}^p [0, d_i]$. That is, associated with each $l \in \ZZ^p$ is
a shifted box $Y_l := \prod_{i=1}^p [l_i d_i, (l_i+1) d_i]$. We call $l \in  \ZZ^p$ the {\em box index} of
$Y_l$. We identify a (non-empty) set of feasible boxes in terms of their box indices $L_f \subset \ZZ^p$. 
The feasible boxes arise from control specifications including obstacle avoidance, collision avoidance, communication 
constraints, and others. Similarly, we identify a (non-empty) set of goal boxes in terms of their box indices 
$L_g \subset L_f$. 

Now consider an output trajectory $y(\cdot, x_0)$ for some control and initial condition $x_0 \in \RR^n$. We associate 
to $y(\cdot, x_0)$ a discretized trajectory called a {\em run} that records the boxes that the output trajectory visits; 
see \cite{BEL08} for a formal definition. The run is denoted as $y_r := l_1 l_2 \cdots$, where $l_i \in \ZZ^p$ is a box
index. We define the {\em behavior} induced by $y(\cdot, x_0)$ to be the sequence of box index increments 
$y_b := \kappa_1 \kappa_2 \cdots$, where $\kappa_i := l_{i+1} - l_i$. Because $y_r$ records every box visited by the output 
trajectory, we have that $\kappa_i \in \{-1,0,1\}^p \setminus \{ 0 \}$. That is, the increment in any coordinate is of 
magnitude at most $1$, and the overall increment is never $0$. Let $\cB$ denote the empty sequence and all finite 
and infinite sequences on $\{-1,0,1\}^p  \setminus \{ 0 \}$. We define a {\em behavioral constraint} $\widehat{\cB} \subset \cB$ 
to be any non-empty subset of $\cB$.  

\begin{problem}[Behavior-Constrained Reach-Avoid]
\label{prob:reachavoid} 
Consider the system \eqref{eq:thesystem} with a gridded output space in terms of grid length vector $d \in \RR^p$. 
We are given goal boxes $L_g$, feasible boxes $L_f$, and a behavioral constraint $\widehat{\cB} \subset \cB$.
Find a feedback control $u(x)$ and a set of initial conditions $\cX_0 \subset \RR^n$ such that for each 
$x_0 \in \cX_0$: 
\begin{itemize}
\item[(i)]
{\bf Avoid}: 
$y(t,x_0) \not \in \RR^p \setminus \left( \bigcup_{l \in L_f} Y_l \right)$ for all $t \geq 0$,
\item[(ii)]
{\bf Reach}:
there exists $T \geq 0$ such that for all $t \geq T$, $y(t,x_0) \in \bigcup_{l \in L_g} Y_l$,
\item[(iii)]
{\bf Behavior}:
$y_b \in \widehat{\cB}$.
\end{itemize}
\end{problem}
 
\section{Hierarchical Motion Primitives}
\label{sec:hierprims}

In this section we present the hierarchical construction of motion primitives. First we recall the definition of 
a maneuver automaton (MA) from our previous work \cite{VUK17, VUK18}, as it serves as the level $0$ or base case 
of the construction.

\subsection{Maneuver Automaton}

A {\em maneuver automaton} (MA) is a hybrid system whose discrete modes correspond to motion primitives. 
Each level $0$ motion primitive has associated to it a continuous time closed-loop vector field obtained by 
applying a state feedback law to \eqref{eq:thesystem}. The edges of the MA model feasible successive motion 
primitives. A motion primitive generates some desired behavior of the output trajectories of the closed-loop 
system over a box in the output space. Because of the uniform gridding of the output space into boxes and 
because of the symmetry assumption on the outputs, level $0$ motion primitives are designed only over $\Ys$. 
The MA applies state resets to ensure that output trajectories remain in $\Ys$. Real, physical trajectories 
clearly do not undergo such resets and instead output trajectories move continuously from box to box. 

\begin{defn} 
\label{defn:H0}
Consider the system \eqref{eq:thesystem} and the box $\Ys$ with grid length vector $d$. 
The level $0$ {\em maneuver automaton} (0-MA) is a tuple
$\cH^0 = ( Q^0, \Sigma^0,  E^0, I^0, Q^{0,0}, X^0, G^0, R^0 )$, where

\begin{description}

\item[State Space]
$Q^0 = M^0 \times \cX^0$ is the hybrid state space, where $M^0 = \{ m_1^0, m_2^0, \ldots \} $ is a 
finite set of level $0$ motion primitives, and $\cX^0 = \RR^n$ is the continuous state space. 

\item[Labels]
$\Sigma^0 = \{ (0, \kappa) ~|~  \kappa \in \{-1,  0, 1\}^p \}$ is a finite set of event labels, where $\sigma = (0, \kappa) \in \Sigma^0$ describes a possible neighbouring direction $\kappa$ from the $p$-dimensional box $\Ys$ at location 0.

\begin{comment} where the first
element $0$ of the pair $( 0, \kappa )$ identifies the starting box in terms of its box index in the gridded 
output space. Since all level $0$ motion primitives are only defined on the box $\Ys$, this index is always $0$.
The second element $\kappa$ is an offset associated with a face of $\Ys$. That is, 
$\kappa = (\kappa_1, \ldots, \kappa_p)$ identifies the face of $\Ys$ given by 
\begin{eqnarray*}
\cF_{\kappa} = 
\left \lbrace 
y \in \Ys ~\middle |~
\begin{cases}
y_i = 0,   & \text{~if~} \kappa_i = -1 \\
y_i = d_i, & \text{~if~} \kappa_i = 1
\end{cases}
\right \rbrace \,.
\end{eqnarray*}
\end{comment}

\item[Edges]
$E^0 \subset M^0 \times \Sigma^0 \times M^0$, is a finite set of edges, describing allowable concatenations of motion primitives.

%An edge $e = ( m, \sigma, m' ) \in E^0$ with label $\sigma = (0,\kappa) \in \Sigma^0$ and $\kappa \in \{ -1, 0, 1\}^p$ denotes that the starting box is $\Ys$ using motion primitive $m \in M^0$, and the next box would be $Y_{\kappa}$ using motion primitive $m' \in M^0$ (modulo a state reset back to $\Ys$). 

\item[Invariants]
$I^0 : M^0 \rightarrow \cP(\RR^n)$ assigns a bounded invariant set $I^0(m)$ to each $m \in M^0$ 
that defines the region on which the vector field $X^0(m)$ is defined. 

\item[Initial Conditions]
$Q^{0,0} \subset Q^0$ is a set of initial conditions. 
%Specifically, $Q^{0,0} = \{ (m,x) \in Q^0 ~|~ x \in I^0(m), m \in M^0 \}$. 

\item[Vector Field]
$X^0 : M^0 \rightarrow \{ f_m^0 \}_{m \in M^0}$ assigns to each $m \in M^0$ a globally Lipschitz closed-loop vector 
field of the form $f_m^0(\cdot) = f( \cdot, u_m(\cdot) )$, where $u_m(\cdot)$ is the state feedback controller 
associated with $m \in M^0$.

\item[Enabling Conditions]
$G^0 : E^0 \rightarrow \{ g_e^0 \}_{e \in E^0}$ assigns to each edge $e = (m, \sigma, m') \in E^0$ a non-empty 
enabling or guard condition $g_e^0 \subset I^0(m)$. 
%We require that the projection of $g_e^0$ into the output space 
%lies in the face of $\Ys$ determined by the label $\sigma$. That is, $h(g_e^0) \subset \cF_{\sigma}$. 

\item[Reset Conditions]
$R^0: E^0 \rightarrow \{ r_e^0 \}_{e \in E^0}$ assigns to each edge $e = ( m, \sigma, m' ) \in E^0$
a reset map $r_e^0 : \RR^n \rightarrow \RR^n$. Resets of states are determined by the event $\sigma \in \Sigma^0$, 
and they only affect the output coordinates in order to maintain output trajectories inside the canonical box $\Ys$. 
%In particular, if $\sigma = (0,\kappa)$, then the reset map is defined so that the $i$-th output component is reset to the right face of $\Ys$ if $\kappa_i = -1$, reset to the left face if $\kappa_i = 1$, and unchanged otherwise.
\hfill \tqed
\end{description}
\end{defn}

The semantics of a level $0$ MA involve the standard notion of an {\em execution} of a hybrid system 
\cite{LYG99,VUK18}.  Informally, the continuous state evolves over the invariant of the current discrete
mode $m \in M^0$ according to the vector field assigned to $m$, and an edge is taken when the continuous 
state reaches an enabling condition, thereby updating the discrete mode as well as resetting the continuous 
state. We shall denote a level $0$ execution as $\chi^0 = (\tau^0, m^0, x^0)$, where $\tau^0$ is a 
{\em hybrid time domain} consisting of time intervals (e.g. $\tau^0 = \{ [0, 2], [2, 3.5], \ldots \}$), 
$m^0$ is a sequence of level $0$ motion primitives recording the motion primitive over each time interval, 
and $x^0$ is the continuous state as a function of time. 

\subsection{Higher Level Maneuver Automata}

We want to extend the notions of motion primitives, maneuver automata, and executions of an MA from level 
$0$ to higher levels by a hierarchical construction that builds level $k$ using only information from level $k-1$. 
We begin with the discrete part of $\cH^0$ since its extension to level $k > 0$ is straightforward. 

The discrete part of $\cH^0$ is given by the tuple $(M^0, \Sigma^0, E^0)$, representing a graph with nodes $M^0$ (the level $0$ motion primitives), 
edges $E^0$, and labels $\Sigma^0$. Analogously, the discrete part of the level $k > 0$ MA, $\cH^k$, is 
a tuple $(M^k, \Sigma^k, E^k)$, representing a graph over level $k$ motion primitives. 
Because level $k$ motion primitives may be defined over more boxes than just $\Ys$, the event
labels $\Sigma^k$ on transitions must be accordingly generalized. 

\begin{comment}
For example, consider the level $1$ motion primitive in Figure~\ref{fig:motiveex} called {\em TwoRight}. 
It is constructed by concatenating twice the level $0$ motion primitive {\em Right}. {\em TwoRight} may, in turn, be concatenated with other level $1$ motion primitives according to the edges $E^1$.
Consider Figure \ref{fig:frames}, where $\mu' \in M^1$ is TwoRight, $\sigma_4 \in \Sigma^1$ is the event denoting completion of TwoRight, and $e = (\mu', \sigma_4, \mu'') \in E^1$ is an edge for some next primitive $\mu''$. First we attach a reference frame in
the gridded output space for the primitive $\mu'$ with an arbitrarily chosen base point $o_{\mu'}$. The role of $\sigma_4 = (l, \kappa) \in \Sigma^1 \subset \ZZ^2 \times \{ -1, 0, 1 \}^2$ is then 
to encode the fact that the transition from $\mu'$ to $\mu''$ occurs moving right from the right box $Y_l$ of TwoRight, which is at position $(0,-2)$ w.r.t. $o_{\mu'}$. Therefore 
$\sigma_4 = (l,\kappa) = ((0,-2),(1,0))$.  In general, if $( m, s, m' ) \in E^k$ is an edge at level $k$,
and $s = ( l, \kappa )$, then $l$ refers to a box index w.r.t. the frame for $m \in M^k$. 
\end{comment}

For example, consider the level $1$ motion primitive in Figure~\ref{fig:motiveex} called {\em U-Turn}. 
It is constructed by concatenating three level $0$ motion primitives: {\em Up}, {\em Right}, and 
{\em Down}. {\em U-Turn} may, in turn, be concatenated with other level $1$ motion primitives encoded by 
an edge $e = ( m, s, m' ) \in E^1$ where $m \in M^1$ denotes {\em U-Turn} and $m' \in M^1$ denotes 
a possible next level $1$ motion primitive. First we attach a reference frame in
the gridded output space for primitive $m$ with an arbitrary base point $o_m$, say the lower left box. The role of $s = (l, \kappa) \in \Sigma^1 \subset \ZZ^2 \times \{ -1, 0, 1 \}^2$ is then 
to encode the fact that the transition from $m$ to $m'$ occurs downwards from the upper right box $Y_l$. Therefore $s = ((1,1),(0,-1))$. 
%In general, if $( m, s, m' ) \in E^k$ is an edge at level $k$, and $s = ( l, \kappa )$, then $l$ refers to a box index w.r.t. the frame for $m \in M^k$. 

Now consider the continuous time part of $\cH^0$. It consists of a continuous state space $\cX^0 = \RR^n$, a
set of continuous time closed-loop vector fields $X^0$, a set of invariants $I^0$ that specify the domain of 
each vector field, and the enabling conditions $G^0$ and reset maps $R^0$ that specify how continuous states 
must be reset upon reaching an enabling condition. These notions are extended to level $k$ as follows.

The continuous state space $\cX^0$ is reinterpreted at level $k$ as a (discrete) state space 
$\cX^k = \ZZ^p \times M^{k-1}$, consisting of the pairs $(l,\mu)$, with $l \in \ZZ^p$ identifying a box index (or offset) and $\mu \in M^{k-1}$ identifying a level $k-1$ 
motion primitive. Similarly, the notion at level $0$ of a continuous state 
trajectory, which is a function of time, is replaced at level $k$ by a discrete sequence of $(l,\mu)$ pairs. 

The continuous time vector fields at level $0$ are replaced by a set of discrete maps. For motion primitive $m \in M^k$, 
the discrete map $f^k_m : \cX^k \times \Sigma^{k-1} \rightarrow \cX^k$ defines the {\em internal transitions} 
on $(l,\mu) \in \ZZ^p \times M^{k-1}$  pairs that make up motion primitive $m$. These discrete maps must adhere 
to the constraints on successive level $k-1$ motion primitives as specified in $\cH^{k-1}$ 
(formal details are below). The invariant $I^k(m)$ is the set of $(l,\mu)$ pairs that the discrete map $f^k_m$ acts on.
The index $l$ in the pair $(l,\mu)$ identifies the origin of the frame for $\mu \in M^{k-1}$ w.r.t. the 
frame for $m \in M^k$. 

The enabling conditions and reset maps take the analogous meanings as in $\cH^0$. They define the 
{\em external transitions} from primitive $m \in M^k$ to other level $k$ motion primitives (including $m$ itself). 
To properly define reset maps, it is necessary to keep track of reference frames for the various primitives involved. 
%To this end, we employ some standard notation from robotics \cite{SPONG}. 
Suppose we have three frames for 
the three primitives $m , m' \in M^k$, and $\mu \in M^{k-1}$, respectively. When an internal or external transition occurs, 
the next primitive effectively starts from a shifted frame. This shift is recorded in terms of the base point of 
each frame, denoted $o_m$, $o_{m'}$, and $o_{\mu}$, respectively. For example, the origin of the frame for 
$\mu$ w.r.t. the frame for $m$ is denoted by $o_{\mu}^m$. Similarly, the origin of the frame for $m'$ 
w.r.t. the frame for $m$ is denoted by $o_{m'}^m$.
% In this notation $o_m^m = o_{\mu}^{\mu} = o_{m'}^{m'} = 0$.
%Note that these shifts of frames of two primitives are not fixed quantities but depend on the particular internal or external transition. 

\begin{defn} 
\label{defn:hmak}

Suppose we are given a level $k-1$~MA $\cH^{k-1}$, where $k \geq 1$. A {\em MA} at level $k$ is a tuple
$\cH^k = (Q^k, \Sigma^k,  E^k, I^k, Q^{k,0}, X^k, G^k, R^k)$, where
\begin{description}

\item[State Space]
$Q^k = M^k \times \cX^k$ is the hybrid state space, where $M^k = \{ m_1^k, m_2^k, \ldots \}$ is a finite set of 
motion primitives at level $k$, and $\cX^k = \ZZ^p \times M^{k-1}$ is the analogue of the notion of a continuous 
state space at level $0$. 
%The state space $\cX^k$ consists of all $(l,\mu)$ pairs where $l$ is the origin of the coordinate frame for motion primitive $\mu \in M^{k-1}$ w.r.t. the frame for some primitive $m \in M^k$. 

\item[Labels]
$\Sigma^k = \{ s_1^k, s_2^k, \ldots, ~|~ s_i^k \in \ZZ^p \times \{-1,  0, 1\}^p \}$ is a finite set of 
event labels. 
If $s = ( l_s, \kappa_s ) \in \Sigma^k$, then $l_s \in \ZZ^p$ is the box index w.r.t. a frame for some primitive $m \in M^k$ from which an external transition occurs, and $\kappa_s$ identifies the face of the box through which the transition occurs. 

\item[Edges]
$E^k \subset M^k \times \Sigma^k \times M^k$ is a finite set of edges describing which level $k$ motion 
primitives can be concatenated. 
%Given $m \in M^k$, the set of transitions associated with edges of the form $e = (m, s, m')$ for some $m' \in M^k$ are called the {\em external transitions} of $m$. 

\item[Invariants]
$I^k : M^k \rightarrow \cP(\cX^k)$ assigns to each motion primitive $m \in M^k$ a non-empty, finite set of 
states $I^k(m) \subset \cX^k$ on which the transition function 
$f^k_m ~:~ \cX^k \times \Sigma^{k-1} \rightarrow \cX^k$ (defined below) acts. 
The box indices $l \in \ZZ^p$ for pairs $(l,\mu) \in I^k(m)$ correspond to the origin of the frame for
$\mu \in M^{k-1}$ w.r.t. the frame for $m \in M^k$. The invariant is different from the {\em envelope} 
of $m \in M^k$, denoted $L^k(m)$, which is the collection of all box indices w.r.t. the frame for $m$ accumulated
from the envelopes of the lower level primitives that constitute $m$. 

\item[Initial Conditions]
$Q^{k,0} \subset Q^k$ assigns a set of initial states, satisfying: if $(m,x) \in Q^{k,0}$, then $x \in I^k(m)$.

\item[Transition Functions]
$X^k : M^k \rightarrow \{f^k_m \}_{m \in M^k}$ assigns to each $m \in M^k$ a transition function 
$f^k_m : \cX^k \times \Sigma^{k-1} \rightarrow \cX^k$ that specifies all the allowable {\em internal transitions} between
$(l,\mu) \in I^k(m)$ pairs that together constitute the level $k$ motion primitive. 
%The transitions defined by $f^k_m$ are called the {\em internal transitions} of primitive $m \in M^k$. 
Consider $(l, \mu) \in I^k(m)$ and $\sigma = ( l_{\sigma}, \kappa_{\sigma} ) \in \Sigma^{k-1}$, such
that $l + l_{\sigma} + \kappa_{\sigma} \in L^k(m)$ (identifying that this is an internal transition) 
and such that $( \mu, \sigma, \mu') \in E^{k-1}$ for some $\mu' \in M^{k-1}$. If
$f_m^k( (l,\mu), \sigma ) = (l',\mu')$, then 
~(i)  $(l',\mu') \in I^k(m)$; 
~(ii) $( \mu, \sigma, \mu') \in E^{k-1}$; and
~(iii) $l = o_{\mu}^m$, $l' = o_{\mu'}^m$, and $l' - l = o_{\mu'}^{\mu}$.

%Referring to Figure~\ref{fig:frames}, (iii) defines the relative positions of the frames for $\mu$ and $\mu'$ with regard to this internal transition. 

\item[Enabling Conditions]
$G^k : E^k \rightarrow \{ g_e^k \}_{e \in E^k}$ assigns to each edge $e = (m,s,m') \in E^k$ a non-empty enabling condition $g_e^k \subset I^k(m)$. If $s = (l_s, \kappa_s) \in \Sigma^k$, the first requirement is that  $l_s + \kappa_s \not \in L^k(m)$, identifying that this is an external transition. Then $g_e^k$ consists of all
those pairs $(l,\mu) \in I^k(m)$ for which an {\em external transition} to a consecutive level 
$k$ motion primitive can occur. That is, $(l,\mu) \in g_e^k$ if there exists (a unique) 
$\sigma = (l_{\sigma}, \kappa_{\sigma}) \in \Sigma^{k-1}$ such that
~(i) there exists $\tilde{\mu} \in M^{k-1}$ such that $(\mu, \sigma, \tilde{\mu}) \in E^{k-1}$;
~(ii)  $l = o_{\mu}^m$, $l_s = l_s^m$ is the index w.r.t. the frame for $m$ for the box from which the external
transition occurs, and $l_{\sigma} = l_{\sigma}^{\mu}$ is the index w.r.t. the
frame for $\mu$ for the same box. Therefore $l_s = l_s^m = o_{\mu}^m + l_s^{\mu} = l + l_{\sigma}$ and 
$\kappa_{\sigma} = \kappa_s$.

\item[Reset Conditions]
$R^k : E^k \rightarrow \{ r_e^k \}_{e \in E^k}$ assigns to each edge $e = (m, s, m' ) \in E^k$ a reset map 
$r_e^k : \cX^k \times \Sigma^{k-1} \rightarrow \cX^k$ that characterizes the {\em external transitions} of $m \in M^k$. 
Associated with $e$ is a unique transformation $o_{m'}^m \in \ZZ^p$ relating the frames of $m$ and $m'$ for
this external transition. 
Consider $s = (l_s, \kappa_s) \in \Sigma^k$, $\sigma = (l_{\sigma}, \kappa_{\sigma}) \in \Sigma^{k-1}$,
and $(l,\mu) \in g_e^k$ such that $l_{\sigma} = l_{\sigma}^{\mu} = l_{\sigma}^m - o_{\mu}^m = l_s - l$ 
and $\kappa_{\sigma} = \kappa_s$. If $r_e^k( (l,\mu), \sigma ) = ( l', \mu' )$, then 
~(i)   $(l',\mu') \in I^k(m')$; 
~(ii)  $( \mu, \sigma, \mu') \in E^{k-1}$; 
~(iii) $l = o_{\mu}^m$, $l' = o_{\mu'}^{m'}$, and $(l' + o_{m'}^m) - l = o_{\mu'}^{\mu}$.
\hfill \tqed
\end{description}
\end{defn}

\begin{figure}%[t]
\centering%
\includegraphics[width=0.75\linewidth,trim=3cm 1cm 1cm 1cm, clip=true]{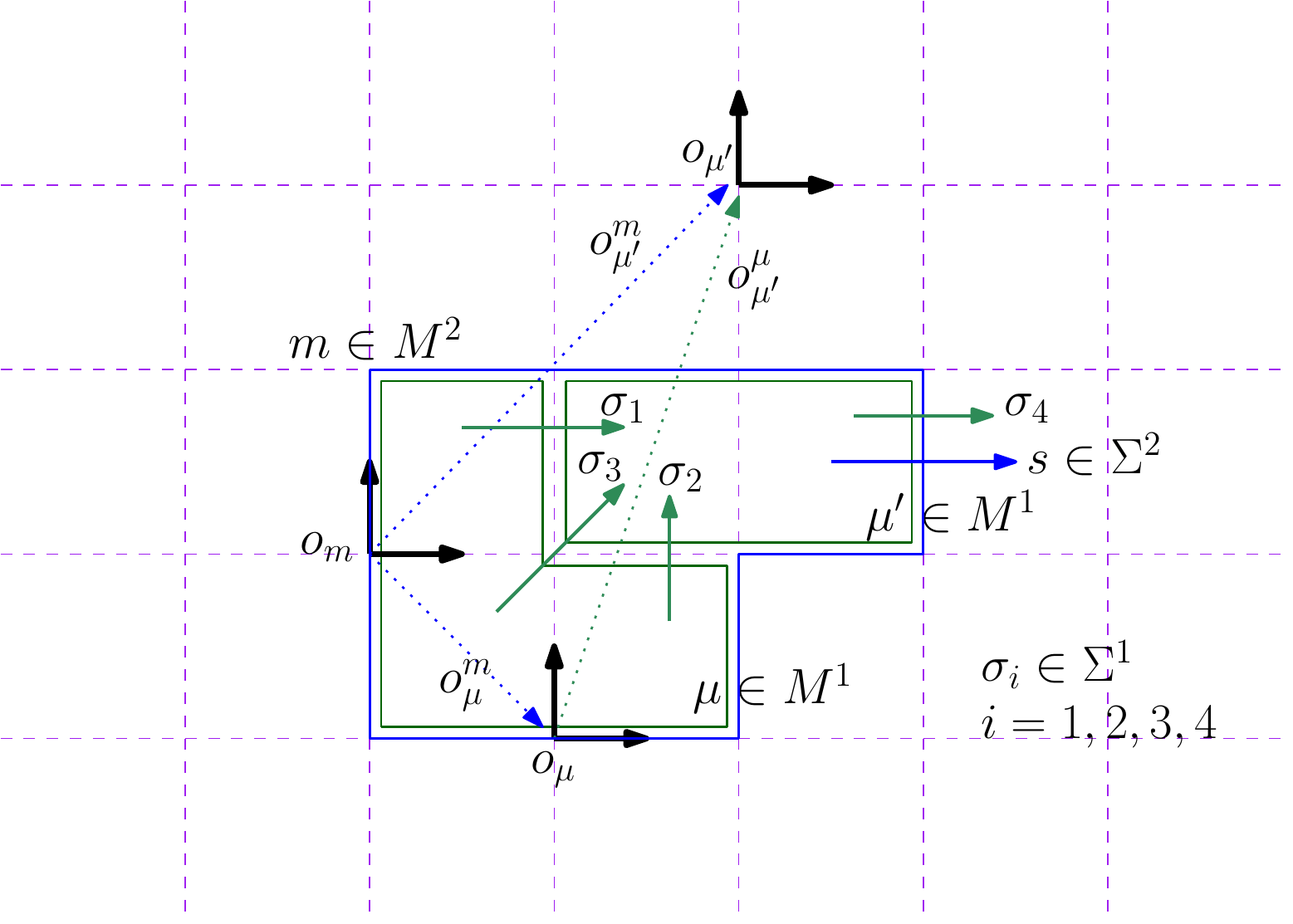}
\caption{An example of level 1 and level 2 motion primitive reference frames, inspired from the 
$p = 2$ output example from Figure \ref{fig:motiveex}. The primitive $m \in M^2$ consists of two 
primitives $\mu, \mu' \in M^{1}$.} 
\vspace{-2mm}
\label{fig:frames}%
\end{figure}

\begin{example}
Returning to the two level hierarchy of Figure~\ref{fig:motiveex}, suppose we want to design motion primitives at level 2 based on the existing motion primitives at level 1. Consider the candidate motion primitive $m \in M^2$ formed by the concatenation of the two level $1$ motion primitives $\mu = Right \& Up$ and $\mu' = Two Right$. Figure~\ref{fig:frames} shows all the geometric information to understand how $\mu$ and $\mu'$ are laid out relative to each other to construct $m$. 

The invariant for $m$ is $I^2(m) = \{ (l,\mu) , (l',\mu') \}$, where $l = (-1,1)$ and $l' = (2,2)$ 
are the indices w.r.t. to the frame for $m$ for the boxes to which the frames for $\mu$ and $\mu'$, respectively, are attached. 
Three internal transitions are encoded in the transition function $f_m^{2}$, corresponding to the level $1$ edges $e_i = (\mu, \sigma_i, \mu') \in E^1$, $i = 1,2,3$. 
For example, the level 2 internal transition $f_m^2((l,\mu),\sigma_1) = (l', \mu')$ is implemented by a 
level $1$ edge $e_1 \in E^1$ with associated event $\sigma_1 = (l_{\sigma_1}, \kappa_{\sigma_1}) \in \Sigma^1$. In the frame of $o_{\mu}$, the transition occurs from the box $l_{\sigma_1} = (-1,1)$ through its 
right face $\kappa_{\sigma_1} = (1,0)$.

There is one level $2$ external transition for $m \in M^2$ with an associated edge $e = (m, s, m') \in E^2$.
In turn, it corresponds to a level $1$ edge $e_4 = (\mu', \sigma_4, \mu'') \in E^1$ and event 
$\sigma_4 \in \Sigma^1$. Thus, $r_e^2((l',\mu'),\sigma_4) = (l'',\mu'')$ for some $(l'',\mu'') \in I^2(m')$. 
The event $\sigma_4 = (l_{\sigma_4}, \kappa_{\sigma_4})$ is expressed in the frame of $o_{\mu'}$, 
while the event $s = (l_s, \kappa_s) \in \Sigma^2$ is expressed in the frame of $o_m$. 
Thus, $l_{\sigma_4} = (0,-2)$, $\kappa_{\sigma_4} = (1,0)$, $l_s = (2,0)$, and $\kappa_s = (1,0)$.
\hfill \tqed
\end{example}

For $k \geq 1$ we define the notation $\cH^{k-1} \preceq \cH^k$ to mean that $\cH^k$ is an abstraction built up from 
$\cH^{k-1}$ according to Definition~\ref{defn:hmak}. In analogy with $\cH^0$, we define an {\em execution at 
level $k > 0$}, denoted as $\chi^k = (\tau^k, m^k, x^k)$. There are two differences from $\chi^0$.
First, the hybrid time domain $\tau^k$ is a sequence of sets of discrete times (e.g. $\tau^k = \{ \{0, 1, 2 \}, \{3\}, \{4,5\}, 
\ldots \}$), where each set consists of the discrete times when internal transitions occur. Second, continuous and 
discrete transitions at level $0$ are replaced by internal and external transitions, respectively, at level $k$.

\subsection{Hierarchical Maneuver Automaton}

Now consider a collection of maneuver automata, $\cH = \{ \cH^k \}_{k=0}^K$, where $K \geq 0$, such that for all 
$k = 1, \ldots, K$, $\cH^{k-1} \preceq \cH^k$. We call $\cH$ a {\em hierarchical maneuver automaton} (HMA).

Next, we can also define an overall {\em hierarchical execution} by stacking together each execution at level $k$.
Suppose that a level $0$ event occurs from the current level $0$ motion primitive. 
This event is either interpreted as an internal or external transition at level $1$. If it is internal, the 
transition function $X^1$ determines the next motion primitive at level $0$ (and the continuous state is reset 
with $R^0$). Otherwise, this event is propagated up the hierarchy until it is registered as internal at some 
level $1 \leq K' \leq K$ or external at level $K$. In the former case, the transition function $X^{K'}$ 
determines the next motion primitive at level $K'-1$, while in the latter case the reset map $R^{K}$ determines 
the next motion primitive at level $K$. An internal transition at level $K'$ is always implemented as an external transition
at the levels $K'-1$ to $0$. The reset maps update the motion primitive at each level below. For example, 
if there is an internal transition at level $3$, then at level $2$, $R^2$ determines the next level $1$ primitive, 
$R^1$ determines the next level $0$ primitive, and $R^0$ resets the continuous state. 

Recalling our definition of the behavior induced by an output trajectory $y(\cdot,x_0)$, we can likewise define
the behavior induced by a hierarchical execution. 

\begin{defn}
Let $\cH = \{ \cH^k \}_{k=0}^K$ be an HMA and let $\chi = \{ \chi^k \}_{k = 0}^K$ be a hierarchical execution of $\cH$. 
The {\em behavior} of $\chi$ is defined to be the behavior of its associated $\chi^0$; namely it is a sequence
$\kappa_1 \kappa_2 \ldots$ where $\sigma_i = ( 0, \kappa_i )$ is the $i$-th event at level $0$.   The {\em language} of 
$\cH$ denoted by $\cL(\cH)$ is the set of all behaviors induced by the hierarchical executions of $\cH$. 
\hfill \tqed
\end{defn}

\section{Solution of Problem~\ref{prob:reachavoid}} 

In this section we give the main elements of the solution to Problem~\ref{prob:reachavoid} using hierarchical 
motion primitives. 

\begin{defn} 
\label{def:icnb}
Let $\cH = \{ \cH^k \}_{k=0}^K$ be a HMA. We say that $\cH$ is {\em well-posed} if (i) each hierarchical 
execution $\chi$ induces a unique behavior in $\cB$, and (ii) for all initial conditions of a hierarchical execution, there exists a maximal extension to a hierarchical execution such that its level 0 execution is infinite.
\hfill \tqed
\end{defn}

Ideally, we would provide a collection of checkable conditions in terms of HMA parameters, and then {\em prove}
that we obtain a well-posed HMA. See \cite{VUK18} where such proofs are supplied 
for $\cH^0$ only. We have adopted Definition~\ref{def:icnb} in order to focus this paper on high level ideas. 

\begin{defn}
Let $\cH = \{ \cH^k \}_{k=0}^K$ be a HMA with $K \geq 1$. 
We say that $\cH$ is {\em complete} if the following hold:
\small
\begin{equation*} 
\label{eq:KHMActrpol}
M^K = \{ m^{\star} \}, \; E^K = \emptyset, \; Q^{K,0} = \{ (m,x) \in Q^K ~|~ x \in I^k(m) \}.
\end{equation*}
\normalsize
That is, at level $K$ there is only one motion primitive, there are no external transitions, and all states are 
valid initial conditions. 
\hfill \tqed
\end{defn}

Completeness provides a hierarchy $\cH$ whose top level acts as a {\em control policy} dictating the assignment of motion 
primitives one level below, with those motion primitives dictating the assignment of motion primitives below that, and so on. In practice, the control policy yielding a complete $\cH$ is obtained by running a planning algorithm at level 
$K-1$ using only the discrete part $ (M^{K-1}, \Sigma^{K-1}, E^{K-1})$, which is a 
(non-deterministic) graph with labeled transitions. For the reach-avoid problem, one may adapt our algorithm from 
\cite{VUK18} for planning at level $K-1$, or other graph search methods may be employed. 

There are two main advantages associated with planning at higher levels. First, a policy produced at a higher level would be much more efficiently computed and represented, composed of only a few high level motion primitives and the transitions between them. Second, the application of planning algorithms to generate policies at higher levels would not need modifications to enforce behavioral constraints explicitly, provided that the concatenation of the constituent high level motion primitives automatically preserve the desired behavioral constraint.

Our main result, Theorem \ref{thm:hierresult}, states conditions to solve Problem~\ref{prob:reachavoid}. We can see that conditions (i), (ii), and (iii) correspond to the avoid, reach, and behavior specifications, respectively. The significance of the result is that if a planning algorithm can meet these conditions, then they satisfy the requirements of Problem~\ref{prob:reachavoid} at the lowest, continuous time level.

\begin{theorem} \label{thm:hierresult}
Consider the system \eqref{eq:thesystem} with a gridded output space in terms of grid length vector $d \in \RR^p$. 
We are given a feasible and goal set of boxes, $L_g \subset L_f \subset \ZZ^p$ and a behavior constraint 
$\widehat{\cB} \subset \cB$. Consider a HMA $\{\cH^k\}_{k = 0}^K$, which is well-posed and complete. Suppose that
\begin{itemize}
\item[(i)] $L^K(m^{\star}) \subset L_f$,
\item[(ii)] all hierarchical executions of $\cH$ reach some subset of boxes of $L_g$ and never leave,
\item[(iii)] $\cL(\cH) \subset \widehat{\cB}$.
\end{itemize}
Then there exists a set of initial conditions $\cX_0$ and an associated feedback control strategy $u(x)$ solving 
Problem~\ref{prob:reachavoid}. 
\end{theorem}

\section{Formation Control} 
\label{sec:formation}

\begin{figure}%[t]
\centering%
\includegraphics[width=1\linewidth,trim=0cm 6.3cm 3cm 0cm, clip=true]{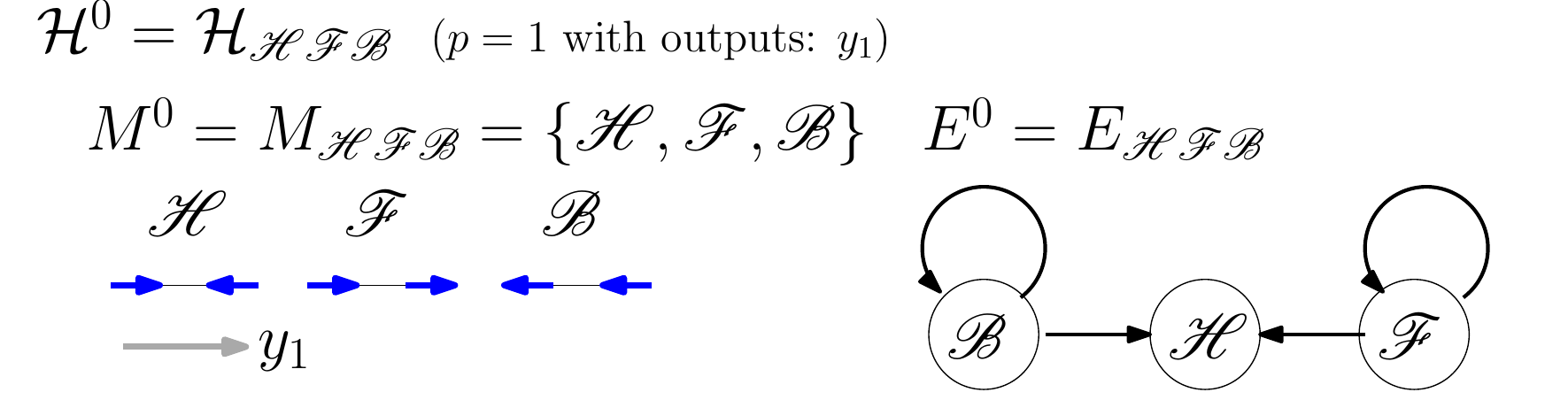}
\includegraphics[width=1\linewidth,trim=0cm 2cm 0cm 0cm, clip=true]{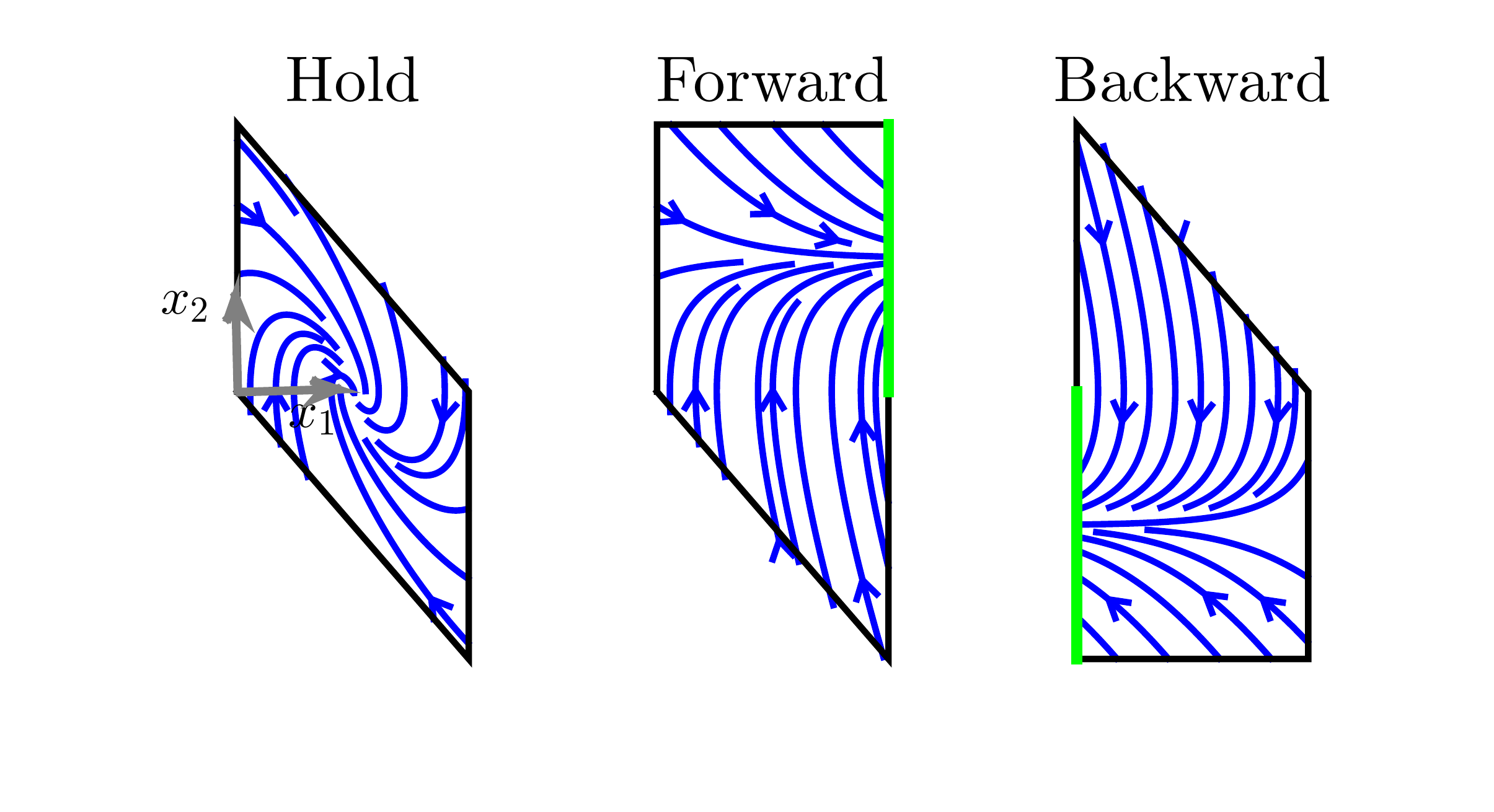}
\vspace{-4mm}
\includegraphics[width=1\linewidth,trim=0cm 0cm 3cm 4.5cm, clip=true]{figs/comp_ex.pdf}
\caption{This figure illustrates a level 0 MA design $\cH^0$ for a system of $p$ double integrators. The top part shows the atomic motion primitives $M^0$ and their edges $E^0$ for one double integrator ($p=1$), which are {\em Hold} ($\sH$), {\em Forward} ($\sF$), and {\em Backward} ($\sB$). Also shown are the associated invariants, vector fields, and enabling conditions (green) in the position-velocity $(x_1,x_2)$ state space. The bottom part shows that two copies of the $p=1$ design may be composed to yield motion primitives for $p = 2$. This procedure extends easily to an arbitrary number of $p$ outputs.} 
\label{fig:comp_ex}%
\end{figure}

\begin{figure}%[t]
\centering%
\includegraphics[width=0.8\linewidth,trim=0cm 0cm 0cm 4.3cm, clip=true]{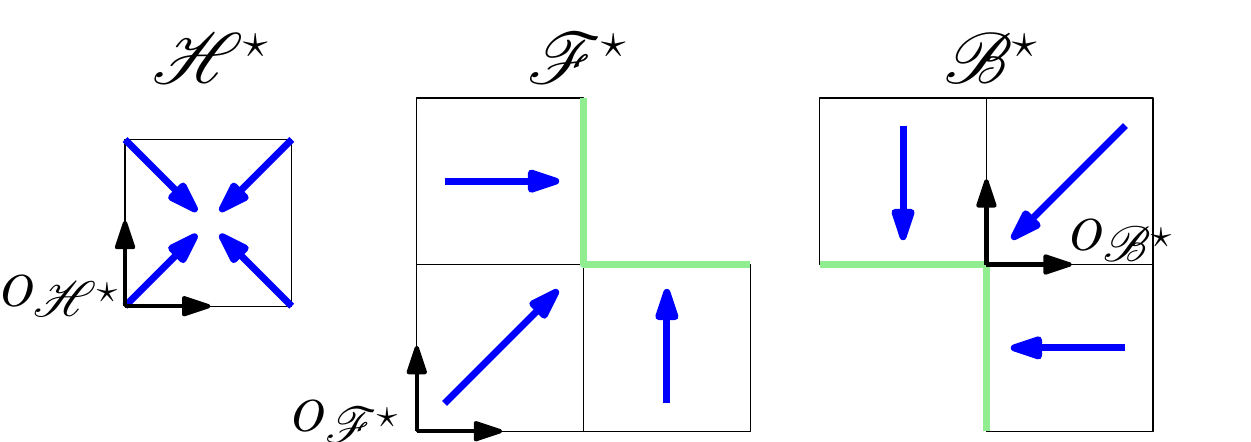}
\caption{This figure illustrates a level 1 MA design $\cH^1$ to achieve the formation behavior constraint. It is built as an abstraction of the level 0 MA $\cH^0$ shown in Figure \ref{fig:comp_ex}. 
%The top row shows the level 1 motion primitives $M^1$ and their edges $E^1$, which are {\em Formation Forward} ($\sFs$), {\em Formation Backward} ($\sBs$), and {\em Formation Hold} ($\sHs$). The bottom row shows the implementation over $\cH^0$ when $p = 2$. 
Although this figure illustrates the case of two outputs $p=2$, this design is easily scalable with respect to any $\cH^0$ (of the form shown in Figure \ref{fig:comp_ex}), which may have any number of $p$ outputs. } 
%\vspace{2mm}
\label{fig:abstr_design}%
\end{figure}

In this section we show how hierarchical motion primitives can be used to solve a formation control problem. 
To illustrate ideas, we consider two agents, each modeled as a double integrator with a scalar output. 
%Our concept has been demonstrated experimentally for four quadrotors each with three outputs, showing that our method scales up well relative to the pedagogical example of this section. 
The behavioral constraint to maintain a formation of two agents, each with one output, is that if one agent incurs an 
increment in its output value due to the application of a motion primitive, then the other agent experiences the same 
increment in the successive motion primitive. Consider a behavior $y_b = \kappa^1 \kappa^2 \ldots \in \cB$ of the 
multi-agent system, where $\kappa^i = ( \kappa^i_1, \kappa^i_2 )$. We require that for all $i \ge 1$,
\begin{equation} \label{eq:behaviordiagonal}
\left| \sum_{l = 1}^i (\kappa^l_1 - \kappa^l_2 ) \right| \le  1 \,.
\end{equation}
Under this behavioral constraint, the relative value of the two outputs is maintained while at the same time a sequence of 
motion primitives to achieve a reach-avoid specification is executed by the multi-agent system. The behavioral constraint can be easily generalized to $N$ agents.

To implement the formation behavioral constraint using hierarchical motion primitives, we propose a two level hierarchy 
consisting of the MA's, $\cH^0$ and $\cH^1$. Level 0 is based on our previous design presented in \cite{VUK17,VUK18}, which we now briefly summarize.

First, the level $0$ (atomic) motion primitives for each agent are: {\em Hold} ($\sH$), {\em Forward} ($\sF$), and {\em Backward} 
($\sB$), corresponding to the agent's output holding its value (remaining in the current box of the gridded 1D output space), 
increasing its value (moving to a box to the right), or decreasing its value (moving to a box to the left). 
Each agent $i$ is equipped with a level $0$ maneuver automaton denoted $\cH^{0,i} = \cH_{\sH \sF \sB}$, $i = 1, \ldots, N$,
as shown in Figure~\ref{fig:comp_ex}. 

Second, to obtain a level $0$ maneuver automaton for the two agent system, we take the parallel composition of the
individual MA's to form the MA $\cH^0 = \cH^{0,1} \Arrowvert \cH^{0,2}$.  
Informally, parallel composition is based on the cartesian product to account for the asynchronous possibilities that arise when combining two independent subsystems.
More generally, for $p$ agents we would form $\cH^0 = \Arrowvert_{j=1}^p \cH^{0,j}$. 
%Details on how to construct the parallel composition of maneuver automata are found in \cite{VUK18}. 
Figure~\ref{fig:comp_ex} illustrates some of the level $0$ motion primitives for the two agents, including 
$\sH \sH$, $\sF \sH$, and so forth for all the possible neighboring directions in 2D. 

Finally, the level $1$ motion primitives for the two agent system are: {\em Formation Forward} ($\sFs$), 
{\em Formation Backward} ($\sBs$), and {\em Formation Hold} ($\sHs$). These are depicted in Figure~\ref{fig:abstr_design}. It is easy to see that the concatenation of these level 1 motion primitives satisfies the desired behavior \eqref{eq:behaviordiagonal}. Once again, this design is easily generalized to $N$ agents.

\section{Experimental Results}
\label{sec:experiment}

It is well known that the model of a quadrotor is differentially flat, and so the nonlinear dynamics can be effectively treated as
three independent double integrators for each of the world frame coordinates $(x_w,y_w,z_w)$ \cite{KUM14}. Suppose we 
have a collection of $N$ quadrotors. The corresponding multi-vehicle system \eqref{eq:thesystem} has $p = 3N$ outputs 
with $n = 6N$ states. We use our formation hierarchical motion primitives to solve Problem \ref{prob:reachavoid} with a formation 
behavior constraint.

First, we grid the 3D physical space and identify obstacles. A formation configuration is specified in terms of the relative box offsets between the vehicles to a chosen representative vehicle and a goal box is specified for the representative. Although the full gridded output space for the $N$ agent system is obtained by composing the gridded 3D physical space $N$ times, we do not compute it explicitly because $N$ may be large. Moreover, instead of specifying the feasible boxes $L_f \subset \ZZ^p$, we define virtual obstacles in 3D based the representative vehicle locations, the formation configuration, and physical obstacles. The behavioral constraint $\widehat{\cB}$ on the full gridded space consists of the constraint \eqref{eq:behaviordiagonal} applied to each physical directions $(x_w, y_w, z_w)$ independently on the $N$ vehicles.
%First we grid the 3D physical space and identify obstacles. A formation configuration is specified in terms of the relative box offsets between the vehicles to a chosen reference vehicle and a goal box is specified for the reference. Conceptually, the full gridded output space for the $N$ agent system is obtained by composing the physical space gridding $N$ times. The behavioral constraint $\widehat{\cB}$ on this gridded space consists of the constraint \eqref{eq:behaviordiagonal} applied to each physical directions $(x_w, y_w, z_w)$ independently on the $N$ vehicles. Since $N$ may be very large, we do not explicitly compute the full output space grid nor the joint feasible and goal boxes $L_f \subset \ZZ^p$ and $L_g \subset L_f$ of the reach-avoid objective corresponding to the physical obstacles and leader goal box.

Next, we group together all outputs of the $N$ agents corresponding to each workspace direction $x_w$, $y_w$, and $z_w$, and
to each such group we appropriate one level $1$ MA, $\cH_{\sHs \sFs \sBs}$. 
Then we take the parallel composition of these three MA's for the three workspace directions to obtain $\cH^1 = \Arrowvert_{j=1}^3 \cH_{\sHs \sFs \sBs}$, which has $3^3 = 27$ motion primitives, independent of $N$. The resulting HMA $\cH = \{\cH^0, \cH^1\}$
is guaranteed to generate the correct behavior, $\cL(\cH) \subset \widehat{\cB}$, because the individual components $\cH_{\sHs \sFs \sBs}$ satisfy \eqref{eq:behaviordiagonal}.
%; that is, if the overall behavior in $p = 3N$ outputs is projected onto the $x_w$, $y_w$, and $z_w$ directions, the formation constraint holds in each of the three directions over $N$ outputs independently (and possibly asynchronously). 

Finally, we generate a control policy $\cH^2$ to yield a complete HMA $\cH' = \{\cH^0, \cH^1, \cH^2 \}$, which also preserves the formation.
The control policy can be generated efficiently based on the 27 motion primitives of $\cH^1$ and the box locations of the representative vehicle using any shortest path planning algorithm. More specifically, the computation of the control policy is equivalent to the complexity of planning as with a single agent, and the computation of the virtual obstacles is linear in the number of agents. Thus we satisfy our aim to dramatically improve the overall computation time.
%Finally, the reach-avoid objective is specified in terms of a set of joint goal boxes $L_g \subset \ZZ^p$ and feasible boxes $L_f \subset \ZZ^p$, and is addressed by computing a control policy $\cH^2$ to yield a complete HMA $\cH' = \{\cH^0, \cH^1, \cH^2 \}$.

\begin{figure}%[htb]
\centering
\begin{tabular}{@{}cc@{}}
\includegraphics[width=.48\linewidth,trim=0cm 0cm 3.7cm 0cm, clip=true]{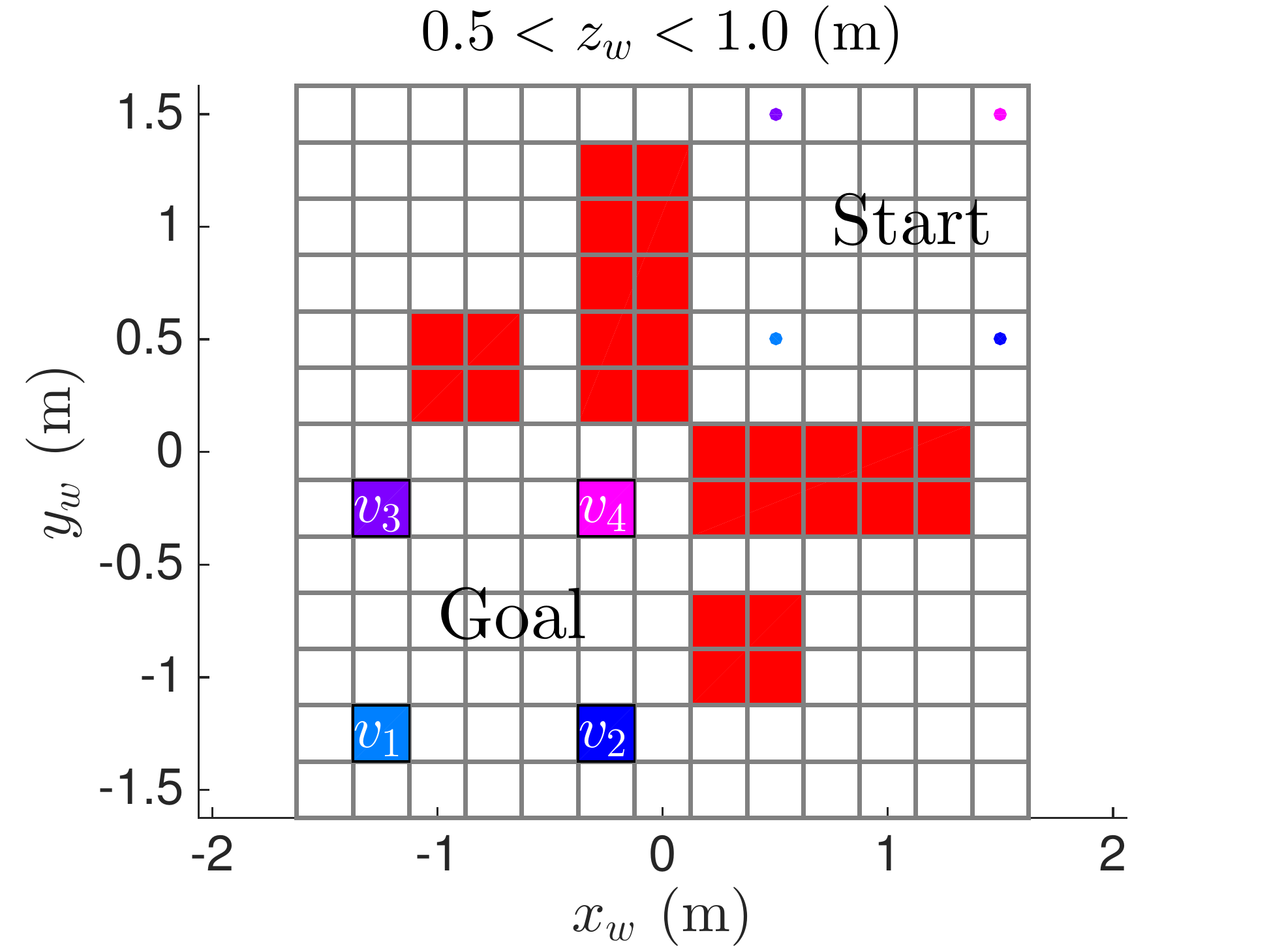} &
\includegraphics[width=.48\linewidth,trim=3.7cm 0cm 0cm 0cm, clip=true]{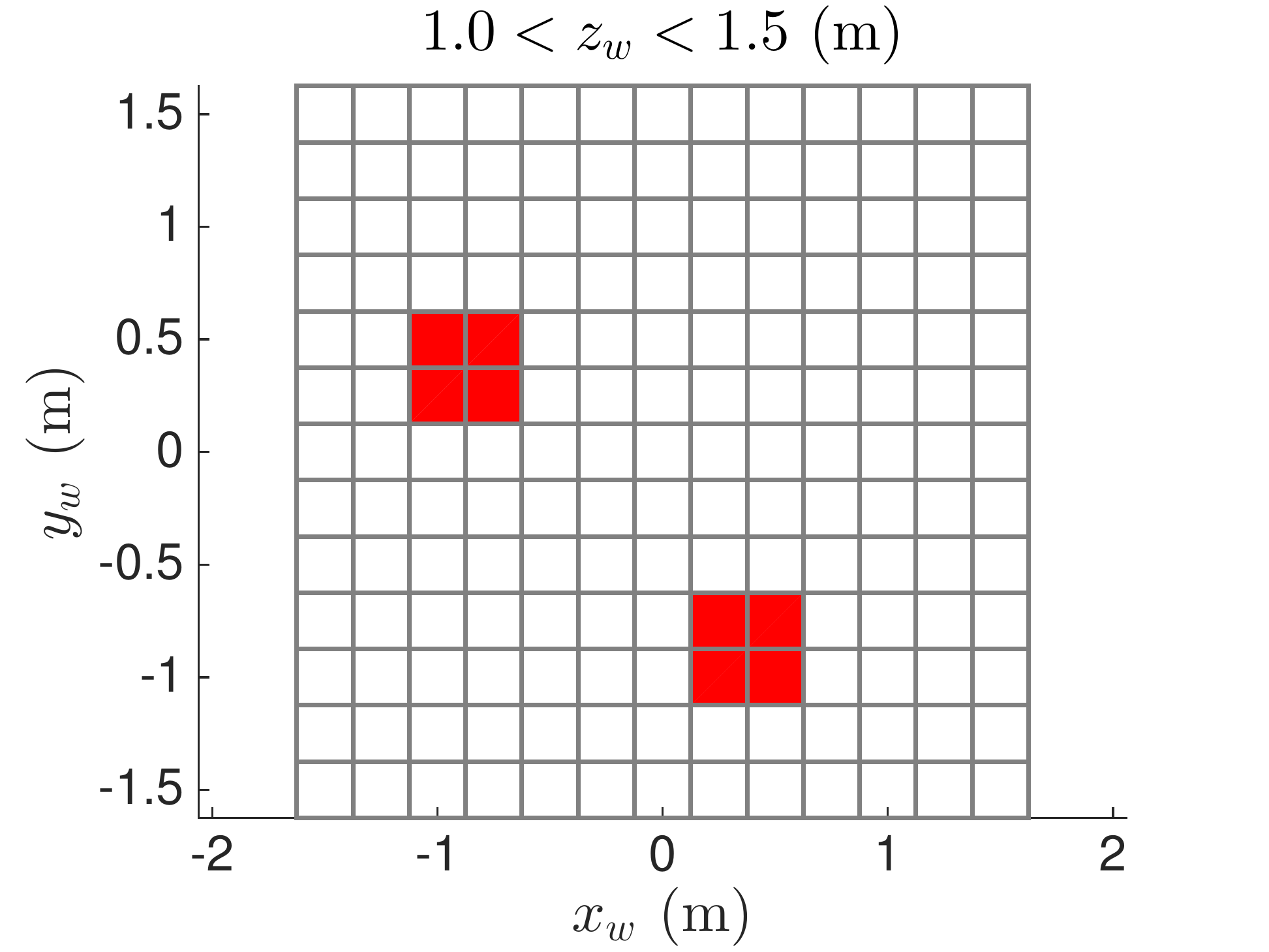} \\
%\multicolumn{2}{c}{\includegraphics[width=0.95\linewidth,trim=0cm 0cm 0cm 4cm, clip=true]{figs/exp_xyz.pdf}} 
\multicolumn{2}{c}{\includegraphics[width=.9\linewidth,trim=3cm 0.5cm 0cm 3cm, clip=true]{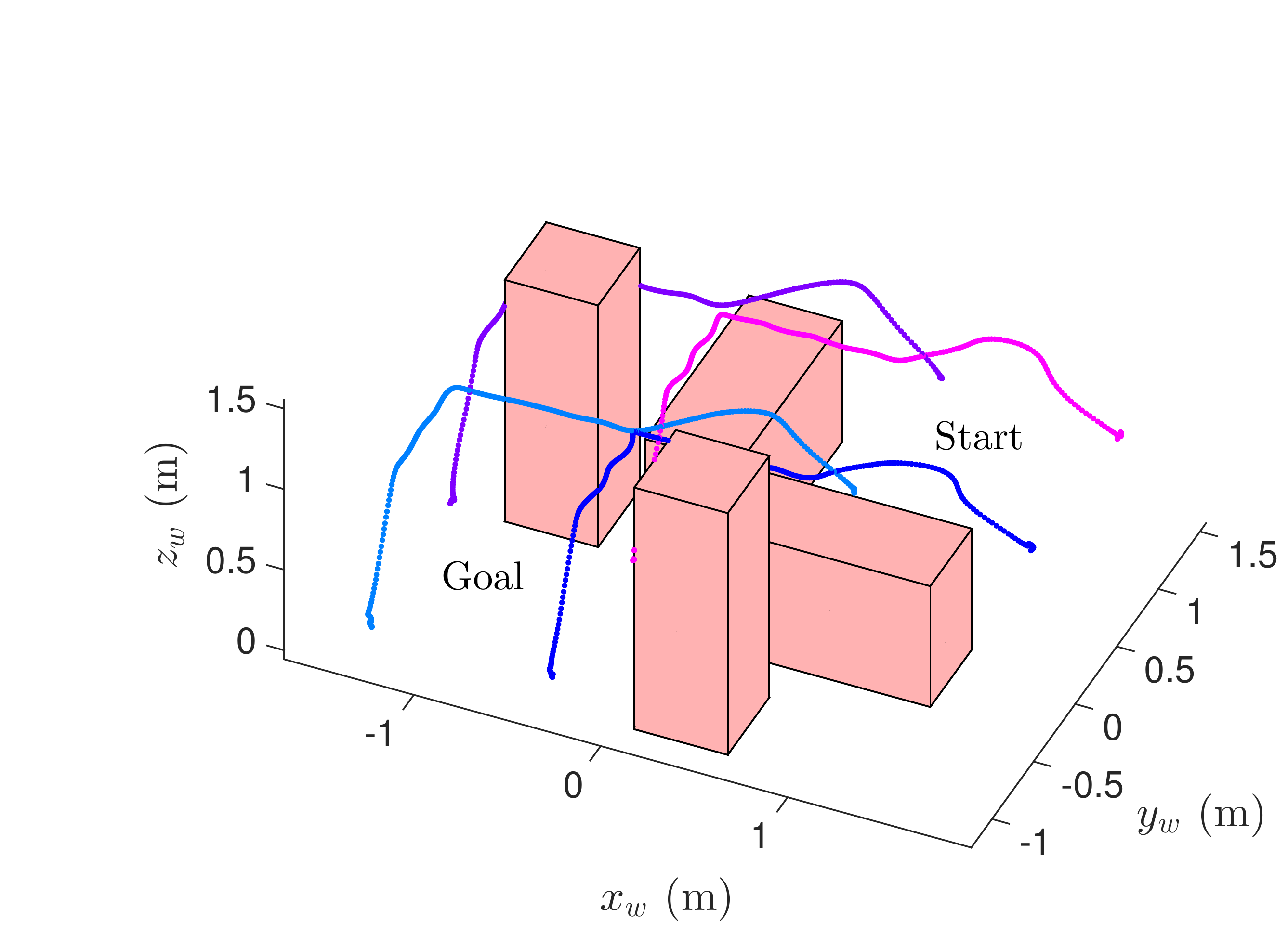}}
\end{tabular}
\caption{Experimental results for the square formation scenario shown in Figure \ref{fig:exp_setup}. }
\vspace{-2mm}
\label{fig:exps}%
\end{figure}

Experimental results are demonstrated for four Crazyflie quadrotors flown indoors using a VICON motion capture system. The top row of Figure \ref{fig:exps} 
shows the gridded 3D workspace, the physical obstacles (red), and the goal boxes (colored). The behavioral 
constraint is imposed to maintain a square formation. This scenario is challenging because of the 
intermingling of obstacles amidst the moving formation and because 3D maneuvers are required for obstacle
avoidance. The middle row of Figure~\ref{fig:exps} shows the 
resulting 3D trajectories in the workspace, demonstrating that the reach-avoid task was successfully executed. 
A video is available at http://tiny.cc/hier-moprim. It also shows different scenarios involving more vehicles in order to demonstrate the applicability to a variety of formations and settings as well as the computational scalability of our 
approach.

\section{Conclusion}
In conclusion, we have introduced hierarchical motion primitives to solve the motion planning problem for a 
large collection of agents. Hierarchical motion primitives are constructed recursively
using a hierarchical maneuver automaton. The framework was applied to multi-vehicle formations, yielding
novel behaviors such as the ability to pass ``through'' obstacles while maintaining a formation. Future work includes developing an automated procedure to generate efficient structures of hierarchical motion primitives for arbitrary reach-avoid behavior-constrained problems and discovering additional applications of our hierarchical approach.

\bibliographystyle{IEEEtranS}

\end{document}